\documentclass{article}

\usepackage{microtype}
\usepackage{graphicx}
\usepackage{subfigure}
\usepackage{booktabs} 

\usepackage{hyperref}

\usepackage[accepted]{icml2025}

\usepackage{amsmath}
\usepackage{amssymb}
\usepackage{mathtools}
\usepackage{amsthm}
\usepackage{graphicx}
\usepackage{booktabs}
\usepackage{multirow} 
\usepackage{multicol}
\usepackage{subcaption}
\usepackage{longtable}
\usepackage{booktabs}
\usepackage{makecell}
\PassOptionsToPackage{svgnames}{xcolor}
\usepackage{colortbl}
\definecolor{mygray}{gray}{.9}
\usepackage[capitalize,noabbrev]{cleveref}

\theoremstyle{plain}

\theoremstyle{definition}

\theoremstyle{remark}

\usepackage[textsize=tiny]{todonotes}

\icmltitlerunning{}

\begin{document}

\twocolumn[
\icmltitle{Multimodal-Guided Dynamic Dataset Pruning for Robust and Efficient Data-Centric Learning}




\begin{icmlauthorlist}
\icmlauthor{Suorong Yang}{nju,ailab}
\icmlauthor{Peijia Li}{nju}
\icmlauthor{Yujie Liu}{ailab}
\icmlauthor{Zhiming Xu}{nju} \\
\icmlauthor{Peng Ye}{ailab,cuhk}
\icmlauthor{Wanli Ouyang}{ailab}
\icmlauthor{Furao Shen}{nju}
\icmlauthor{Dongzhan Zhou}{ailab}
\end{icmlauthorlist}

\icmlaffiliation{nju}{National Key Laboratory for Novel Software Technology, Nanjing University}
\icmlaffiliation{ailab}{Shanghai Artificial Intelligence Laboratory}
\icmlaffiliation{cuhk}{The Chinese University of Hong Kong}
\icmlcorrespondingauthor{Furao Shen}{frshen@nju.edu.cn}
\icmlcorrespondingauthor{Dongzhan Zhou}{zhoudongzhan@pjlab.org.cn}

\icmlkeywords{Machine Learning, ICML}

\vskip 0.3in
]


\printAffiliationsAndNotice{}
\begin{abstract}
Modern deep models are trained on large real-world datasets, where data quality varies and redundancy is common. 
Data-centric approaches such as dataset pruning have shown promise in improving training efficiency and model performance. 
However, most existing methods rely on static heuristics or task-specific metrics, limiting their robustness and generalizability across domains.
In this work, we introduce a dynamic dataset pruning framework that adaptively selects training samples based on both task-driven difficulty and cross-modality semantic consistency.
By incorporating supervision from pretrained multimodal foundation models, our approach captures training dynamics while effectively filtering out uninformative samples.
Our work highlights the potential of integrating cross-modality alignment for robust sample selection, advancing data-centric learning toward more efficient and robust practices across application domains.
\end{abstract}

\section{Introduction}
\label{sec:intro}
Recent progress in AI is largely driven by large training datasets.
Yet, the vast volume of training data poses substantial computational and efficiency challenges, which hinder the data efficiency and accessibility, especially for researchers and startups without large-scale infrastructures~\cite{moderate,dataset_pruning}.
Beyond raw sizes, large-scale datasets often contain redundant and semantically inconsistent samples~\cite{noise-dataset-1,noise-dataset-2,noise-dataset-3}, further inflating training costs without necessarily improving performance.
These limitations have sparked a growing interest in \textbf{data-centric methods} that aim to optimize the \textit{quality} rather than the \textit{quantity} of data.

To mitigate this, existing methods focus on dataset distillation~\cite{dataset_distillation,dataset_distillation2,dataset_condensation3,dataset_distillation4,dataset_distillation6} and static data selection~\cite{moderate,moso,dataset_pruning,yang2023not,glister}, which aim to synthesize or select a smaller fixed subset from the whole datasets before training begins.
While the training costs are reduced due to the decrease in dataset sizes, the distillation and selection algorithms typically require extra costs, which can sometimes exceed the costs of training models on large-scale datasets~\cite{forgetting,tdds,infobatch} (see Table~\ref{tab:imagenet-1k} on ImageNet-1k~\cite{imagenet}).
More critically, these approaches may overlook changes in sample difficulty and the evolving dynamics of the model during online training, potentially leading to suboptimal performance.
In contrast, dynamic dataset pruning~\cite{infobatch,dynamic_pruning} adjusts the data composition along with the training process, continuously selecting samples based on the model's evolving learning needs.
However, most existing pruning and curation approaches are domain-specific and rely on a single modality to prioritize hard samples, lacking robustness to potential noise or ambiguity in practice.
For instance, scoring metrics such as loss values may be unreliable in the presence of noisy samples, potentially leading to inaccurate sample selection decisions.
These challenges highlight the need for a robust, cross-modality, and dynamic dataset pruning method. 

To this end, we propose a dynamic dataset pruning framework with dual-supervision optimization, which integrates task-specific supervision with cross-modality semantic consistency supervision to adaptively curate high-quality training samples throughout training. 
Specifically, task-specific loss is leveraged to serve as a direct indicator of training progress; however, relying solely on loss values can lead to suboptimal selection.
The suboptimal performance arises because noisy or corrupted samples often exhibit high loss, and incorporating them may degrade overall model performance.
To mitigate this issue, we further introduce cross-modality semantic consistency signals derived from a pretrained multimodal model CLIP~\cite{clip}, which captures the alignment between visual and textual modalities. 
The additional supervision allows the pruning mechanism to filter out semantically irrelevant or inconsistent data, promoting the selection of samples that are not only hard but also semantically meaningful.
By combining task-driven and cross-modality criteria, our framework enhances the robustness and generalizability of data pruning across diverse learning scenarios.
For instance, with the existence of noisy data, our framework effectively filters out noisy data points, demonstrating its robustness in real-world data pipelines.
We believe this work offers an efficient and practical solution for data-centric learning, especially in settings that demand both scalability and cross-modality robustness.

\section{Related Work}
Data-efficient learning can be categoried into dataset distillation~\cite{dataset_distillation5,dataset_distillation6,dataset_distillation7,wang2025datasetdistillationneuralcharacteristic}, dataset condensation~\cite{dataset_condensation1,dataset_condensation2,dataset_condensation3}, static data selection~\cite{dataset_pruning,moderate,yang2023not,moso,tdds}, and dynamic data pruning~\cite{infobatch,dynamic_pruning,yang2025dynamicdataselectionmeets}. In contrast to distillation, which synthesizes a minimal training set, data selection and pruning reduce the training costs by selecting original training samples from the entire dataset. 

Static data selection selects a coreset from the full training dataset before the training begins.
Existing methods can be generally categorized into score-based~\cite{data_diet,moso,tdds,cgscore,forgetting,hu2025donod}, dataset distribution-based~\cite{moderate,ccs}, and optimization-based~\cite{dataset_pruning,glister,clip-selection,cgscore}.
Among these, score-based methods compute the sample-wise importance scores, followed by sorting and selecting samples based on the the average of the $\ell_2$-norm error vector~\cite{data_diet}, the expectation of gradient norm~\cite{data_diet}, error-based~\cite{forgetting,score-based-3}, the variation of gradient in training progress~\cite{tdds}, and the change of the optimal empirical risk~\cite{moso}.
Based on the geometric distribution, these methods select samples in terms of sample distance~\cite{herding,moderate}, sample coverage~\cite{ccs,k-center-selection}, and samples' neighboring relationship~\cite{d2}.
Optimization-based methods select samples through various optimization techniques, such as gradient matching~\cite{opt-based-3,core-set}, self-supervised pruning metrics~\cite{beyond}, influence function~\cite{dataset_pruning,influence-func-based}, bi-level optimization~\cite{glister}, and submodularity~\cite{cgscore,opt-based-1,opt-based-4}.

Unlike static selection methods that determine a fixed subset before training, dynamic dataset pruning continuously adjusts sample selection throughout training, reducing unnecessary iterations and improving efficiency.
Existing methods leverage model uncertainty or loss-based heuristics for dynamic selection. 
The work~\cite{dynamic_pruning} proposes UCB and $\epsilon$-greedy strategies to estimate the uncertainty value for each training sample and select a given fraction of training samples with the highest scores.
During each training epoch, models are trained on these selected samples.
The work~\cite{dynamic_pruning-2} employs the dynamic uncertainty as scores to guide selection.
InfoBatch~\cite{infobatch} leverages vanilla loss values to select samples during training.
Based on the instantaneous loss values calculated from current model snapshots, more informative samples are selected during training.
\section{The Proposed Method}
\subsection{Preliminary}
We denote the full training dataset as $\mathcal{D}=\left\{\left.\left(x_i, y_i\right)\right|_{i=1} ^N\right\}$, where $x$ refers to a sample, $y$ to its corresponding label, and $N$ is the total number of samples. 
Unlike static data selection, which fixes a subset before training, dynamic dataset pruning selects a sequence of subsets $\hat{\mathcal{D}}=\left\{ \mathcal{D}_1, \mathcal{D}_2, ..., \mathcal{D}_T \right\}$, one per epoch, based on the model's evolving state, with each $\mathcal{D}_t$ maintaining a target selection ratio.
At each epoch $t$, the model $f_\theta$ is updated on $\mathcal{D}_t$ via:
\begin{equation}
\theta_t = \theta_{t-1}-\eta_t \nabla \mathcal{L}\left(b_t, \theta_{t-1}\right),
\end{equation}
where $\eta_t$ is the learning rate at the $t$-th step, $\nabla \mathcal{L}$ is the gradient of the loss function, and $b_t$ is a mini-batch of the batch size $b_s$ from $\mathcal{D}_t$.

The goal of dynamic dataset pruning at time $t{-}1$ is to select a subset $\mathcal{D}_{t}$ that best matches the model’s current needs:
\begin{equation}
    \mathcal{D}_t=\underset{\mathcal{D}_t \subseteq \mathcal{D}, \|\mathcal{D}_t\|=k}{\arg \min } \mathbb{E}_{z:(\boldsymbol{x}, y) \sim P}\left[\mathcal{L} \left(z, \theta_{t-1}^*\right)\right],
\end{equation}
where $P$ denotes the dataset distribution, $z$ is a test sample sampled from $P$.
Here, $\theta_{t-1}^*$ is the empirical risk minimizer updated using the selected datasets from $\mathcal{D}_0$ to $\mathcal{D}_{t-1}$. 
Selecting $k < N$ samples per epoch reduces training costs by $(N{-}k) \cdot T$ forward passes and approximately $\lceil (N{-}k) \cdot T / b_s \rceil$ backward updates.
 
\subsection{Overview}
We introduce a dynamic dataset pruning framework to achieve efficient, lossless training acceleration by adaptively selecting the most informative and semantically correct samples during online training.
Unlike methods that rely on static and one-shot scoring metrics based on the current model snapshot, our approach optimizes a selection score for each sample.
The selection score dynamically adjusts selected datasets as training progresses and is optimized based on two complementary supervisory signals: task-specific loss and multimodal semantic consistency loss. 
The task-specific loss captures sample difficulty and reflects the current model training progress, while the cross-modality semantic consistency loss filters out noisy samples by ensuring semantic alignment.
This dual-supervision mechanism enhances both informativeness and reliability in sample selection.
By continuously refining sample scores based on evolving training dynamics, our framework provides an adaptive and robust selection process, effectively reducing redundancy while preserving model performance.

\subsection{Training Dynamics-Aware Task Supervision}
Estimating sample importance over the course of training requires supervision signals that reflect the model’s current learning state. To this end, we leverage the loss values of samples as supervision based on the following two reasons: (i) they can be obtained online during training at no extra computational cost;
(ii) they serve as effective and straightforward indicators of model learning status, as samples with relatively high loss often provide more informative patterns for learning~\cite{curriculum,curriculum_learning,entaugment,yang2024adaaugment}.
Thus, we define these loss values as difficulty signals, denoted by $\boldsymbol{s}_\mathcal{T}$.
Under our setting, $\boldsymbol{s}_\mathcal{T}=\mathcal{L}(f_\theta(x_i),y_i)$.
However, high loss values are not always indicative of useful samples—noisy or corrupted data often exhibit similarly high loss~\cite{noisy_labels,noisy_labels-2}, making it difficult to distinguish them from genuinely informative hard examples. Filtering such samples purely through image features is time-consuming and unreliable~\cite{denoise,denoise-2}.
To address this, we introduce cross-modality semantic supervision from a pretrained multimodal foundation model, which assesses whether a sample maintains alignment between visual and textual modalities.

\subsection{Cross-Modality Semantic Supervision}
Real-world datasets inevitably involve label noises and image corruption~\cite{noise-dataset-1,noise-dataset-2,moderate}, which can mislead models during training. 
Therefore, we adopt the pretrained multimodal foundation model, CLIP~\cite{clip}, for computing cross-modality semantic alignment, which can filter out noisy samples and prioritize semantically aligned data.
Moreover, to alleviate the domain shifts and discrepancies between the pretrained and target datasets, we incorporate lightweight dataset-specific adapters~\cite{clip-adapter,tip-adapter} for both image and text encoders.
Both adapters are merely linear layers and fine-tuned using InfoNCE loss~\cite{infonce,infonce2} for knowledge transfer, while the pretrained CLIP weights are frozen. This significantly enhances the zero-shot performance of pretrained multimodal foundation models.

Specifically, let $E_I$ and $E_T$ denote the image and text encoders, respectively, each equipped with its corresponding adapter.
Given an image and its category label, we use the prompt template ``\textit{A photo of [CLASS]}'' to obtain text features $\boldsymbol{z}_t$ using $E_T$.
We then define the scaled cosine similarity as:
\begin{equation}
    sim(\boldsymbol{z}_i, \boldsymbol{z}_t)=\tau \frac{\boldsymbol{z}_i \cdot \boldsymbol{z}_t}{\|\boldsymbol{z}_i\| \|\boldsymbol{z}_t\|},
\end{equation} 
where $\tau$ is the scaling temperature, set to $\log(1 / 0.07)$~\cite{clip}, and the image feature $\boldsymbol{z}_i$ is obtained using $E_I$.
To assess the alignment between visual and textual features, we use the scaled cosine similarity as the cross-modality consistency signal, i.e., $\boldsymbol{s}_\mathcal{C}=sim(\boldsymbol{z}_i, \boldsymbol{z}_t)$.
The higher values of $\boldsymbol{s}_\mathcal{C}$, the stronger the semantic alignment.
To enhance the efficiency, the modality features are precomputed for all samples.
During training, semantic consistency scores are directly computed from feature embeddings, introducing minimal overhead while preserving cross-modality alignment.
As a result, this cross-modality signal complements the task loss supervision, enabling our framework to robustly prune training data by balancing task-driven difficulty with semantic reliability.
\begin{table*}[h]
    \centering
    \captionof{table}{Comparison with state-of-the-art baselines. All methods are trained with ResNet-18 on CIFAR-10/100. Note that some results could not be computed due to the unavailability of open-source code and parameter settings, making it impossible to reproduce. Random* means randomly selecting samples in each epoch.\label{tab:comparison_experiment}}
    \vspace{-1mm}
	\resizebox{.75\textwidth}{!}{
    \begin{tabular}{c|ccc|ccc}
    \bottomrule[1.5pt]
    Dataset &  \multicolumn{3}{c|}{\cellcolor{mygray}CIFAR-10}& \multicolumn{3}{c|}{\cellcolor{mygray}CIFAR-100}   \\ \hline
    Whole Dataset &\multicolumn{3}{c|}{95.6}&\multicolumn{3}{c}{78.2}  \\ \hline
    Selection Ratio (\%)& 30& 50&70& 30& 50&70  \\ \hline
    
    Random &90.2&92.3&93.9&69.7&72.1&73.8  \\ 
    Herding~\cite{herding} &80.1 &88.0&92.2 &69.6&71.8&73.1  \\
    EL2N~\cite{data_diet} &91.6&95.0&95.2 &69.5&72.1&77.2   \\
    GraNd~\cite{data_diet} &91.2&94.6&95.3 &68.8&71.4&74.6   \\
    Glister~\cite{glister} &90.9&94.0&95.2 &70.4&73.2&76.6   \\
    Forgetting~\cite{forgetting} &91.7&94.1&94.7 &69.9&73.1&75.3  \\
    Moderate-DS~\cite{moderate} &91.5&94.1&95.2&70.2&73.4&77.3   \\
    Self-sup. prototypes~\cite{beyond} &91.0&94.0&95.2&70.0&71.7&76.8   \\
    DP~\cite{dataset_pruning}&90.8&93.8&94.9 &-&73.1&77.2   \\
    Random* &93.0&94.5&94.8 &74.4&75.3&77.3   \\
    UCB~\cite{dynamic_pruning} &93.9&94.7&95.3 &-&75.3&77.3   \\
    $\epsilon$-Greedy~\cite{dynamic_pruning} &94.1&94.9&95.2 &-&74.8&76.4  \\
    InfoBatch~\cite{infobatch} &94.7&\textbf{95.1}&95.6 &76.5&78.1&78.2   \\ \hline
    Ours &\textbf{94.9} &\textbf{95.1} &\textbf{95.9} &\textbf{77.4}&\textbf{78.7}& \textbf{78.9}  \\ \bottomrule[1.5pt]
    \end{tabular}}
\end{table*}
\subsection{Dynamic Data Selection Optimization}
Furthermore, rather than directly selecting samples based on $\boldsymbol{s}_\mathcal{T}$ and $\boldsymbol{s}_\mathcal{C}$ values, we optimize a learnable selection score through a lightweight numerical optimization process.
This continuous optimization inherently captures training dynamics, as supervision signals evolve with the model. It reflects how each sample contributes to learning over time, mitigating the risk of over-emphasizing transiently hard examples and promoting a more balanced and stable selection process. As training progresses, this mechanism ensures dynamic adaptation to sample utility and reduces the influence of noisy or outlier samples.

We initialize $\boldsymbol{s}$ uniformly with ones to maintain an unbiased starting point and optimize it using a multi-objective loss:
\begin{equation}\label{eq:loss_s}
    \mathcal{L}_{\boldsymbol{s}} = \frac{1}{\|\boldsymbol{s}\|_0} \boldsymbol{s}\cdot(\lambda \boldsymbol{s}_{\mathcal{C}} - \boldsymbol{s}_{\mathcal{T}}) ,
\end{equation}
where $\lambda$ is a weighting coefficient constant. This loss promotes samples that are semantically aligned and informative, while penalizing those with high task loss but low semantic consistency—often indicative of noise.
Based on $\boldsymbol{s}$, we select samples with scores near the median of $\boldsymbol{s}$, which serves as a robust proxy for distributional balance~\cite{median}. 
Consequently, the optimized selection results smooth out short-term fluctuations in training loss, yielding more stable selection trajectories.

Regarding the parameter complexity of our method, it is proportional to the dataset volume, which is $\mathcal{O}(N)$.
Thus, the parameter complexity is marginal compared to popularly used deep models. For instance, on ImageNet-1k with 1.2 million samples, the parameter complexity increases by less than 5.5\% using ResNet-50.
Moreover, since our method does not rely on any auxiliary model or surrogate network, the computational overhead introduced by the score optimization is negligible relative to the target model training cost.
Thus, this makes our approach both lightweight and scalable for large-scale training scenarios.

\section{Experiment}
In this section, we evaluate the effectiveness of our method under both clean and noisy data scenarios. This setup reflects real-world scenarios where data quality varies and robustness is essential. Our experiments aim to assess whether the proposed cross-modality supervision improves generalization on clean data and maintains strong performance under noise, demonstrating its practicality for scalable, data-centric learning.
\subsection{Comparison with State-of-the-arts}
We evaluate our method’s generalization ability and training efficiency across both medium-scale (CIFAR-10/100) and large-scale (ImageNet-1k) datasets. As shown in Table~\ref{tab:comparison_experiment}, our method consistently outperforms existing baselines—including both static and dynamic selection strategies—achieving the highest accuracy across all benchmarks. Notably, even simple dynamic pruning methods like Random* surpass many advanced static selection methods, highlighting the importance of dynamically adjusting the training set during learning.  

On CIFAR-10/100, our approach achieves higher accuracy than full-data training using only 70\% of the samples, leading to approximately 30\% cost savings. 

\subsection{Generalization on ImageNet-1k}
We show the results on large-scale ImageNet-1k, the results in Table~\ref{tab:imagenet-1k}.
It can be observed that our method scales effectively to large datasets, significantly outperforming static baselines that require heavy offline computations such as surrogate training, Gram matrix inversion, or bi-level optimization~\cite{moderate,clip-selection}. In contrast, our framework performs lightweight online pruning with minimal overhead. At a 60\% selection ratio, it accelerates training by reducing forward-backward operations while simultaneously improving model performance. These results demonstrate the method’s practical value in real-world, resource-constrained scenarios.

\begin{table*}[h]
       \centering
    \caption{Results on ImageNet-1k with a 60\% selection ratio using ResNet-50 on an 8-A100 server. Note that due to the high computational costs and device memory costs~\cite{moderate}, Glister and CG-Score are not reported. Some results are from~\cite{infobatch}. Time is the wall clock time; Overall (n*h)  is the total GPU hour.\label{tab:imagenet-1k} }
    \vspace{-1mm}
	\resizebox{0.95\textwidth}{!}{
    \begin{tabular}{c|cccccccccc|cc}
    \toprule[1.5pt]
    Method &Herding&EL2N&GraNd&Forgetting&SSP&Moderate&UCB&Infobatch&Glister&CG-Score&Ours&Whole Dataset \\ \hline
    Acc. (\%) &71.1&72.3&71.0&72.5&70.0&73.1&75.8&76.5&-&-&\textbf{76.8}&76.4 \\ \hline
    Time (h) &10.5&10.5&10.5&10.5&10.5&10.5&10.5&10.5&10.5&10.5&10.5&17.5 \\
    Overhead (h) &$>$17.5&$>$17.5&$>$17.5&$>$17.5&$>$24.0&$>$17.5&0.03&0.0028&-&-&0.0030&0.0 \\
    Overall (n*h) & $>$224.0 &$>$224.0&$>$224.0&$>$224.0&$>$276.0&$>$224.0&\textbf{84.0}&\textbf{84.0}&-&-&\textbf{84.0}&140.0 \\
    \bottomrule[1.5pt]
    \end{tabular}}
    \vspace{-3mm}
\end{table*}

\subsection{Illustration of the Selection Results}

\begin{figure}[]
    \centering
    \includegraphics[width=1.\linewidth]{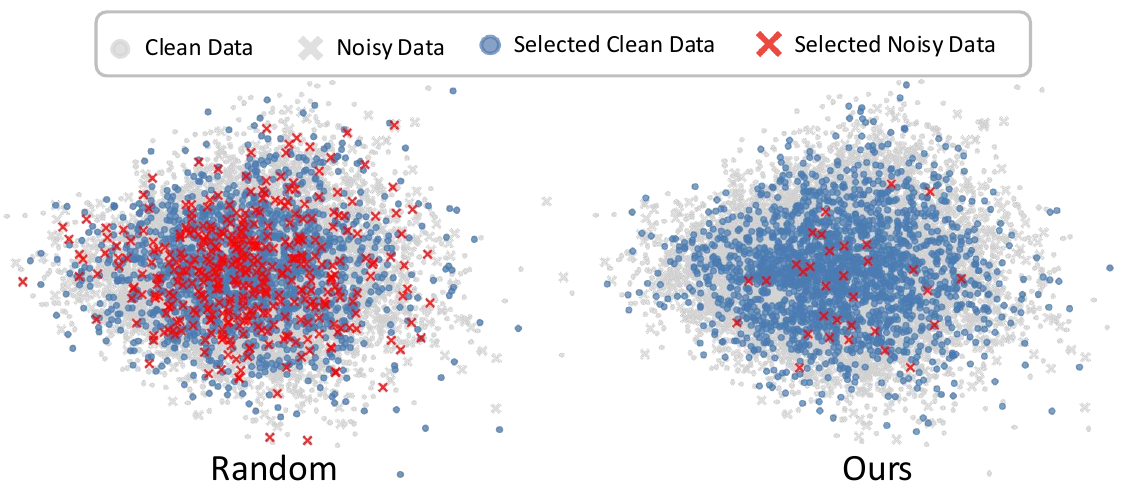}
    \caption{Illustration of the selected data in noisy conditions. The noisy ratio and selection ratio are 20\%.}
    \label{fig:illustration-selection}
\end{figure}

To better understand the robustness of our selection mechanism under data corruption, we visualize the selected samples on datasets with a 20\% label noise ratio and a fixed selection ratio. As shown in Figure~\ref{fig:illustration-selection}, our method consistently filters out noisy or mislabeled data while retaining semantically meaningful samples. This demonstrates its ability to perform reliable data curation even under imperfect supervision, reinforcing its applicability in real-world, noise-prone settings.

\section{Discussion and Future Work}
In this section, we outline several potential limitations and directions for future research.
1) Our method relies on a pretrained CLIP model to provide cross-modality supervision for guiding sample selection. While CLIP has demonstrated strong performance across a wide range of tasks~\cite{Zhou_2022,ramesh2022hierarchicaltextconditionalimagegeneration,wu2023clipself,gu2024rwkvcliprobustvisionlanguagerepresentation}, its effectiveness may be constrained in specialized domains where it lacks relevant pretraining coverage, such as medical imaging, scientific diagrams, or domain-specific modalities. 
Adapting the framework to such domains may require incorporating domain-adapted encoders or alternative alignment signals.
2) The current formulation is tailored for classification tasks with category-level textual prompts. Extending the approach to more complex vision tasks, such as object detection, semantic segmentation, or vision-language retrieval, presents an exciting direction for generalizing the framework to broader data-centric settings.
3) Although the selection optimization is computationally lightweight and scales well in terms of training cost, maintaining a sample-wise score vector introduces $\mathcal{O}(N)$ memory complexity. On extremely large datasets, this can become a bottleneck—a challenge shared by most score-based selection methods. Future work could explore compressed representations or gradient-based proxy signals to further reduce memory overhead without compromising selection quality.

\section{Conclusion}
This study emphasizes the significant advantages of incorporating cross-information into data curation.
We address a key limitation of prior pruning methods by proposing a dynamic dataset pruning framework that adaptively adjusts the training data composition over time, reducing redundancy while accelerating training without sacrificing performance. 
By integrating task-specific loss and semantic consistency supervision from a pretrained multimodal CLIP model, our method offers a concrete recipe for integrating training dynamics and multimodal alignment into sample selection.
We hope our work provides an exploration toward more robust, scalable, and generalizable data-centric learning across modalities and domains.


\bibliography{example_paper}
\bibliographystyle{icml2025}


\end{document}